\documentclass[graybox]{svmult}

\usepackage{type1cm}        
\usepackage{makeidx}         
\usepackage{graphicx}        
\usepackage{multicol}        
\usepackage[bottom]{footmisc}

\usepackage{newtxtext}       %
\usepackage[varvw]{newtxmath}       

\makeindex             

\begin{document}

\title*{Identification, explanation and clinical evaluation of hospital patient subtypes}
\author{Enrico Werner, Jeffrey N. Clark, Ranjeet S. Bhamber, Michael Ambler, Christopher P. Bourdeaux, Alexander Hepburn, Christopher J. McWilliams, Raul Santos-Rodriguez}
\authorrunning{Enrico Werner, Jeffrey N. Clark, Ranjeet S. Bhamber et al.} 
\institute{Enrico Werner \at University of Bristol, Bristol (UK), \newline\email{enrico.werner@bristol.ac.uk}
\and Jeffrey N. Clark \at University of Bristol, Bristol (UK), \newline\email{jeff.clark@bristol.ac.uk} \and Ranjeet S. Bhamber \at University of Bristol, Bristol (UK), \newline\email{ranjeet.bhamber@bristol.ac.uk} 
\and Michael Ambler and Christopher J. McWilliams\at University of Bristol and University Hospitals Bristol NHS Foundation Trust, Bristol (UK), \newline \email{mike.ambler@bristol.ac.uk and chris.mcwilliams@bristol.ac.uk}
\and Christopher P. Bourdeaux \at University Hospitals Bristol NHS Foundation Trust, Bristol (UK), \newline\email{christopher.bourdeaux@uhbristol.nhs.uk}
\and Alexander Hepburn and Raul Santos-Rodriguez \at University of Bristol, Bristol (UK), \newline\email{alex.hepburn@bristol.ac.uk} and enrsr@bristol.ac.uk}
 
%
%
\maketitle


\abstract{We present a pipeline in which unsupervised machine learning techniques are used to automatically identify subtypes of hospital patients admitted between 2017 and 2021 in a large UK teaching hospital. With the use of state-of-the-art explainability techniques, the identified subtypes are interpreted and assigned clinical meaning. In parallel, clinicians assessed intra-cluster similarities and inter-cluster differences of the identified patient subtypes within the context of their clinical knowledge. By confronting the outputs of both automatic and clinician-based explanations, we aim to highlight the mutual benefit of combining machine learning techniques with clinical expertise.}
\abstract*{We present a pipeline in which unsupervised machine learning techniques are used to automatically identify subtypes of hospital patients in a large UK teaching hospital admitted between 2017 and 2021. With the use of state-of-the-art explainability techniques, the identified subtypes are interpreted and assigned clinical meaning. In parallel, clinicians assessed intra-cluster similarities and inter-cluster differences of the identified patient subtypes within the context of their clinical knowledge. By confronting the outputs of both automatic and clinician-based explanations, we aim to highlight the mutual benefit of combining machine learning techniques with clinical expertise.}

\section{Introduction}

Patients admitted to hospital constitute a heterogeneous population with different levels of illness severity, morbidities, response to treatments and outcomes \cite{castela_forte_identifying_2021}. Therefore, predicting the right treatment is challenging even when patients are initially diagnosed with the same conditions. For diagnosis and determining treatment options, physicians rely on factors including the patient's medical history \cite{baytas_patient_2017}, their own clinical experience and their professional intuition \cite{castela_forte_identifying_2021}. 

Advances in computing technologies and the introduction of electrical health records (EHR) mean that more information is available to physicians than ever before. However, hospitals are still in the process of transitioning from paper records to EHR, which leads to challenges when analyzing the data and inferring high-level information \cite{baytas_patient_2017}. As intensive care units (ICUs) are the most data-rich hospital department, machine learning approaches have mostly focused on these environments  \cite{vranas_identifying_2017,castela_forte_identifying_2021,anand_predicting_2018,mcwilliams_towards_2019}. Recent progress has also been made for general wards \cite{carr_evaluation_2021,oei_towards_2021,giannini_machine_2019,cheng_using_2020}.  

Outcome prediction and risk scoring are of high clinical importance. Several risk scoring methods have been developed and deployed, e.g. Rothman index \cite{rothman_development_2013}, MEWS \cite{subbe_validation_2001}, APACHE IV \cite{balkan_evaluating_2018}, and SOFA \cite{khwannimit_comparison_2007}. The National Early Warning Score 2 (NEWS) is increasingly used in UK hospitals \cite{abbott_pre-hospital_2018} and has good predictive ability in patients with infections and sepsis \cite{alam_exploring_2015}. However, for respiratory diseases like COVID-19, the results are conflicting \cite{baker_national_2021,kostakis_performance_2021,carr_evaluation_2021}. This raises the question: are early warning scores such as NEWS equally effective for all patient subtypes? 

We argue that the predictive ability of scores could be further improved by incorporating insightful patient subtyping. Historically, patients were grouped based on their level of sickness, i.e., the creation of ICUs. The reorganization presented an innovation, as expertise in caring for the critically ill could be focused on a single location \cite{costa_organizing_2016}. Patient subtyping could be the next innovative step, namely, instead of focusing solely on the severity of their sickness, patients could be further grouped based on their clinical needs \cite{vranas_identifying_2017}. In a pilot study, non-ICU patients were physically grouped in based on similar patient characteristics rather than diagnoses, leading to a reduced admittance of low-risk patients to ICU from 42\% to 22\%. Additionally, the average ICU length of stay was reduced from 4.6 to 4.1 days \cite{dlugacz_expanding_2002}. 

Automatic patient subtyping aims to assign patients to clinically meaningful groups using factors such as their disease progression, medical history, EHR, and ultimately paves the path to precision or personalised medicine by tailoring diagnostic and therapeutic strategies to the patient's needs \cite{baytas_patient_2017,mirnezami_preparing_2012}. Subtyping can be framed as an unsupervised machine learning task, using clustering methods to identify distinct high-density regions separated by sparse regions within a dataset \cite{davies_cluster_1979}. These clusters represent patients that are in some sense similar according to the data. Clustering algorithms such as k-means and hierarchical clustering have recently been applied to identify clusters in a general ICU population \cite{vranas_identifying_2017}, cardiovascular clusters in sepsis patients \cite{geri_cardiovascular_2019}, and corticosteroid response in patients with severe asthma \cite{wu_multiview_2019}. 

However, clustering alone is insufficient to provide practical support to determine treatment options. The resulting clusters also must be interpreted such that clinicians can validate and learn from cluster assignments. While previous studies manually assigned clinical meaning to the clusters after their creation, before these models can be widely deployed in hospitals, the final users must `trust' the models. This requires an in-depth understanding of the models' behaviour and confidence in individual predictions \cite{ribeiro_why_2016}. Model-agnostic explainability approaches such as LIME and variants \cite{ribeiro_why_2016, sokol2019blimey} can be used for explaining the predictions of clustered data \cite{zhou_model-agnostic_2019}. This paper presents a pipeline in which we:

\begin{itemize}
\item Propose the use of unsupervised machine learning techniques to identify patient subtypes on admission for a new dataset of hospital patients from a large UK teaching hospital. 
\item Implement a combination of explainability techniques and statistical properties of the clusters to evaluate and assign clinical meaning to the identified subtypes. 
\item In parallel and independently, hospital clinicians derive the main clinical properties of the identified subtypes using additional records, a key and necessary step in developing human-in-the-loop machine learning systems in medical settings.
\end{itemize}

\section{Method}

\subsection{Data Source and Conditions}
Subjects are patients admitted to the Bristol Royal Infirmary, a large teaching hospital covering most medical and surgical specialties. The clinical characteristics of this historical data source is summarised in Table \ref{table_full_dataset}. Only patients were considered for which the NEWS and all corresponding vitals were available i.e. temperature, systolic blood pressure, heart rate, hemoglobin saturation with oxygen (SATS), respiratory rate, and level of consciousness. Only vitals taken within the first 24 hours after hospital admission were considered. Patient visits lasting less than 2 hours were considered as routine appointments and omitted. Some patients were admitted several times and each admission was considered as an independent event. Patients with restricted or limited level of consciousness are described as `unconscious'.

\begin{table}[!h]
\begin{tabular}{p{4.5cm}p{3cm}}
Number of patients	& 60,731 \\
Number of admissions & 95,825 \\
Gender (\% female) & 51.1 \\
Age (years) & 58.96 ($\pm$21.13) \\ 
Length of stay (hours) & 47.84 ($\pm$ 107.26)\\
Mortality (\%) & 2.84\\
NEWS & 1.53 ($\pm$1.79) \\
Temperature	(°C) & 36.81 ($\pm$0.54) \\
Systolic blood pressure (mmHg)	& 123.97 ($\pm$21.16) \\
Heart rate (bpm)	& 79.51 ($\pm$15.96) \\
SATS (\%)	& 96.03 ($\pm$2.62) \\
Respiratory rate (bpm)	&  17.43 ($\pm$2.77) \\
Limited level of consciousness (\%)	& 0.84 \\

\end{tabular}
\sidecaption
\caption{Clinical characterization of the full dataset. NEWS = National early warning score, SATS = Hemoglobin saturation with oxygen. Value format is mean (standard deviation).}

\label{table_full_dataset}
\end{table}

\subsection{Dimensionality Reduction and Clustering}
To improve interpretability, dimensionality reduction was performed using Uniform Manifold Approximation and Projection (UMAP) \cite{mcinnes_umap_2018} based on the six vitals: temperature, systolic blood pressure, heart rate, SATS, respiratory rate and level of consciousness. The first three vitals were scaled, the latter three transformed with the logit function. After dimensionality reduction, HDBScan \cite{campello_density-based_2013} was applied to the embedding to identify clusters subsequently used for clinical validation. 


\subsection{Clustering Explanations}
\label{ch_cluster_explanation}

Understanding how and why patients were clustered was a fundamental requirement to establish clinical trust and ultimately reduce the risk of unintended harm. Each vital sign's contribution for the assignment of patients into each cluster was determined by generating and mutating 25,000 surrogate samples per cluster using the TabularBlimeyTree decision tree explainer \cite{sokol2019blimey} within \texttt{FAT Forensics} (v0.1.1): an open source toolbox \cite{sokol_fat_2022}\footnote{\url{https://github.com/fat-forensics/fat-forensics}}.  in order to identify decision boundaries and determine each vital's contribution. Input features were the scaled vital signs. The probabilistic argument was set to False. Default arguments and settings were otherwise used. All generated samples were visualised in the embedding space to ensure reasonableness.

\subsection{Clinical Evaluation}
\begin{figure*}[b]
    \centering
    \includegraphics[width=\textwidth]   
    {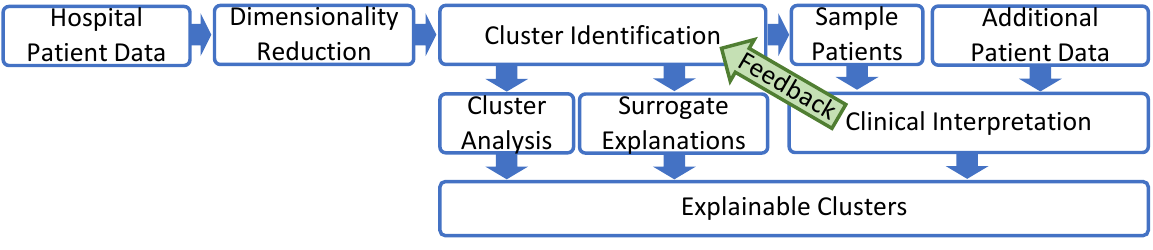}
    \caption{Pipeline overview, from dataset import to output of explainable clusters.}
    \label{Flowchart}
\end{figure*}

Clinical validation was conducted, independently of the previously described analysis, by providing two clinicians with ten sample patients for each cluster. These patients were selected uniformly at random and visually checked to ensure consistency with the underlying representation for each cluster. The clinicians then pulled additional information of these patients from the hospital databases and analysed their clinical records to assess and evaluate intra-cluster similarities and inter-cluster differences according to both the data and their clinical knowledge (Figure \ref{Flowchart}).    

\begin{table*}[b]
\centering
\caption{Clinical characterization of all individual clusters. SATS = Hemoglobin saturation with oxygen. Value format is mean (standard deviation). ICD10 (10th revision of International Classification of Diseases) specifies codes for diseases and diagnoses, where for each cluster the most frequent code corresponds to the following; Cluster 0 sepsis, Cluster 1 abdominal and pelvic pain, Cluster 2, 3 \& 4 chronic ischemic heart disease.}
\begin{tabular}{p{3.0 cm}p{1.6cm}p{1.6cm}p{1.6cm}p{1.6cm}p{1.6cm}}
& \textbf{Cluster 0}	& \textbf{Cluster 1}	& \textbf{Cluster 2}	& \textbf{Cluster 3}	& \textbf{Cluster 4} \\
\hline
Number of patients & 434 & 7583 & \textbf{41641} & 9050 & 13752 \\
Number of admissions	& 453	& 8713	& \textbf{61022}	& 10080	& 15557 \\
Gender (\% female) & 47.7 & \textbf{63.6} & 48.3 & 57.3 & 51.1 \\
Age (years) & \textbf{69.9 \newline ($\pm$ 18.0)} & 49.6 \newline($\pm$ 21.6) & 64.1 \newline($\pm$ 18.9) & 52.3 \newline($\pm$ 21.6) & 56.2 \newline($\pm$ 21.0) \\ 
Length of stay (hours) &  \textbf{74.5 \newline($\pm$ 162.8)} & 40.5 \newline($\pm$ 102.5)& 52.2 \newline($\pm$ 112.6) & 38.1 \newline($\pm$ 90.6) & 40.2 \newline($\pm$ 94.6)\\
Mortality	(\%) & \textbf{21} &	 2	& 4	& 1 & 1 \\
ICD10 code \newline(most frequent) & A41.9	& R10.3	& I251	& I251	& I251 \\
NEWS & \textbf{5.78 \newline($\pm$ 2.78)} & 0.99 \newline($\pm$1.31)	& 1.65 \newline($\pm$1.88) & 0.78 \newline($\pm$1.12) & 0.69 \newline($\pm$1.03) \\
Temperature	(°C)& 36.68 \newline($\pm$0.68)	& 36.79 \newline($\pm$0.56)	& \textbf{36.85 \newline($\pm$0.62)}	& 36.75 \newline($\pm$0.46)	& 36.73 \newline($\pm$0.44) \\
Systolic blood pressure \newline(mmHg) & 121 \newline($\pm$26.33)	& 126 \newline($\pm$21.09)	& \textbf{130 \newline($\pm$23.13)}	& 128 \newline($\pm$21.19)	& 129 \newline($\pm$20.54) \\
Heart rate	(bpm)& 79.18 \newline($\pm$15.86) & 78.10 \newline($\pm$16.23)	& \textbf{81.62 \newline($\pm$17.53)}	& 77.04 \newline($\pm$14.91)	& 76.54 \newline($\pm$13.82) \\
SATS	(\%)&  95.72 \newline($\pm$3.13)	& \textbf{100.00 \newline($\pm$0.02)} & 95.53 \newline($\pm$2.02)	& 99.00 \newline(0)	& 98.00 \newline(0.09) \\
Respiratory rate	(bpm)& \textbf{18.14 \newline($\pm$4.60)}	& 16.72 \newline($\pm$2.47)	& 17.53 \newline($\pm$2.95)	& 16.63 \newline($\pm$2.11)	& 16.63 \newline($\pm$2.06) \\
Limited level of \newline consciousness (\%)& \textbf{100} & 0.18 &	 0.12 	& 0.06	& 0.03 \\
\end{tabular}
\label{table1}
\end{table*}
\noindent

\section{Results}

\subsection{Cluster Characterization}
The data extracted between Nov 2017 to March 2021 comprised 116,004 cases (70,452 patients). Of these, 95,825 cases (60,731 patients) had all
\begin{figure}[!ht]
    \centering
    \includegraphics[height=0.85\textheight]{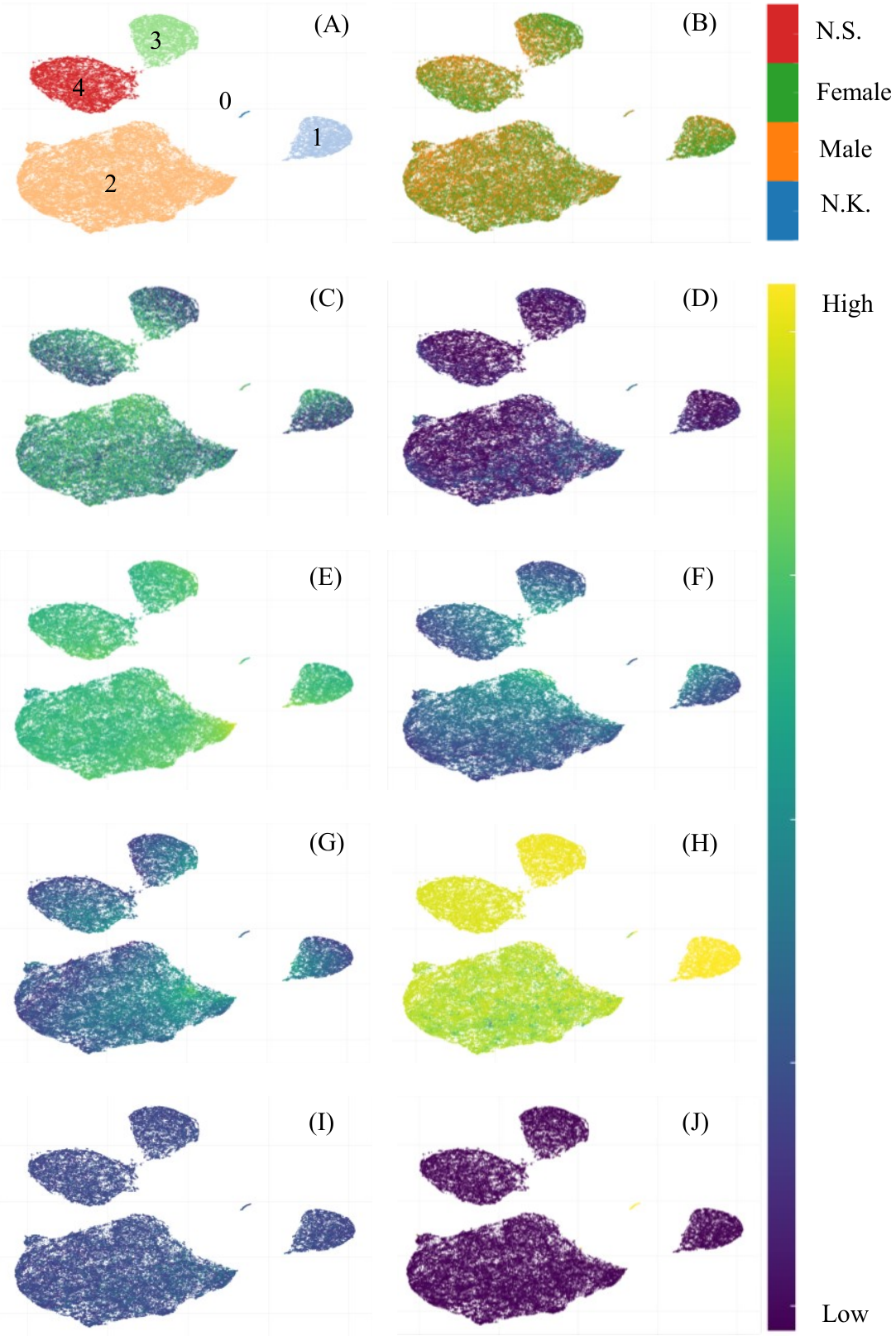}
    \caption{Parameters mapped onto the embedding space after dimensionality reduction. Cluster assignment (A), gender (B) where N.K. = not known, N.S. = not specified, age (C), NEWS (D), temperature (E), systolic blood pressure (F), heart rate (G), hemoglobin saturation with oxygen (H), respiratory rate (I), level of consciousness where high = limited level of consciousness (J)}
    \label{Clusters}
\end{figure}
 vitals taken within the first 24 hours of their $\geq$2 hours hospital stay and were included in the study.

Dimensionality reduction and clustering revealed five clusters (Figure \ref{Clusters}A, summarised in table \ref{table1}). Most vitals plus gender and age reveal a gradient in values across or within clusters. Globally, the NEWS (Figure \ref{Clusters}D) is highest for cluster 0 but also shows high values in some areas in cluster 2. Temperature (Figure \ref{Clusters}E) reveals a gradient for all clusters, most noticeable in cluster 2. Systolic blood pressure (BP, Figure \ref{Clusters}F) behaves similarly. However, the increase in BP seems to be directed towards the unoccupied region between cluster 2, 3, and 4. Heart rate (Figure \ref{Clusters}G) reveals a gradient within all clusters. In contrast, SATS (Figure \ref{Clusters}H) has a more global gradient with values increasing from cluster 2 to cluster 4 to cluster 3 and 1. The intra-cluster SATS values are homogeneous within clusters 1, 3, and 4. Cluster 2 has a moderate gradient, and cluster 0 has a strong gradient. 

Respiratory rate (Figure \ref{Clusters}I) is almost homogeneous with only moderate variations within all clusters. The level of consciousness is an exception (Figure \ref{Clusters}J). Cluster 0, the smallest cluster, contains almost all patients with limited level of consciousness and no fully "conscious" patients (Table \ref{table1}). Only a small number of patients limited level of consciousness can be found distributed in the other clusters. Gender (Figure \ref{Clusters}B) and age (Figure \ref{Clusters}C) appear to show similarly distributed gradients, with female patients tending to be admitted at a lower age.

\begin{figure}[b]
    \centering
    \includegraphics[width=\textwidth]
    {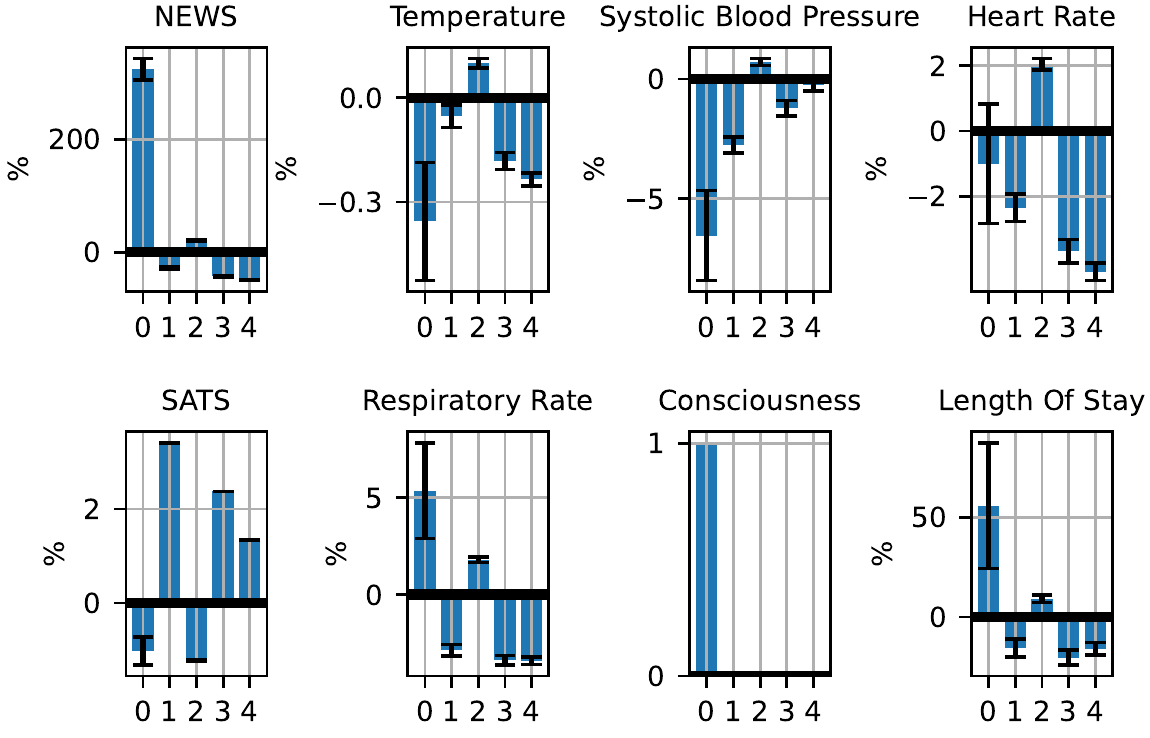}
    \caption{Vitals, NEWS and length of stay for individual clusters. The mean value of each cluster is compared to the mean value (black line) of the whole population. Error bars represent 95\% confidence intervals.}
    \label{Vitals_barplot_confidence}
\end{figure}
\subsection{Cluster Visualisation and Vitals}
A goal of analysing the individual clusters is to identify their unique characteristics relative to each other and the overall population. Figure \ref{Vitals_barplot_confidence} shows the percentage difference in measurements for each cluster relative to the overall population.

Cluster 0 shows the clearest difference compared to all other clusters with higher than average NEWS, respiratory rate, consciousness, and longest hospital stay, and a decreased  temperature and systolic blood pressure. The high consciousness score indicates that cluster 0 contains almost all patients with limited level of consciousness. Additionally, cluster 0's confidence interval covers the largest range, indicating higher variability. For clusters 1-4, the confidence intervals all have similar magnitudes. Features of cluster 2 appear distinct as they are directionally opposite to clusters 1, 3, and 4. 

\begin{figure*}[b]
    \includegraphics[width=\textwidth]
    {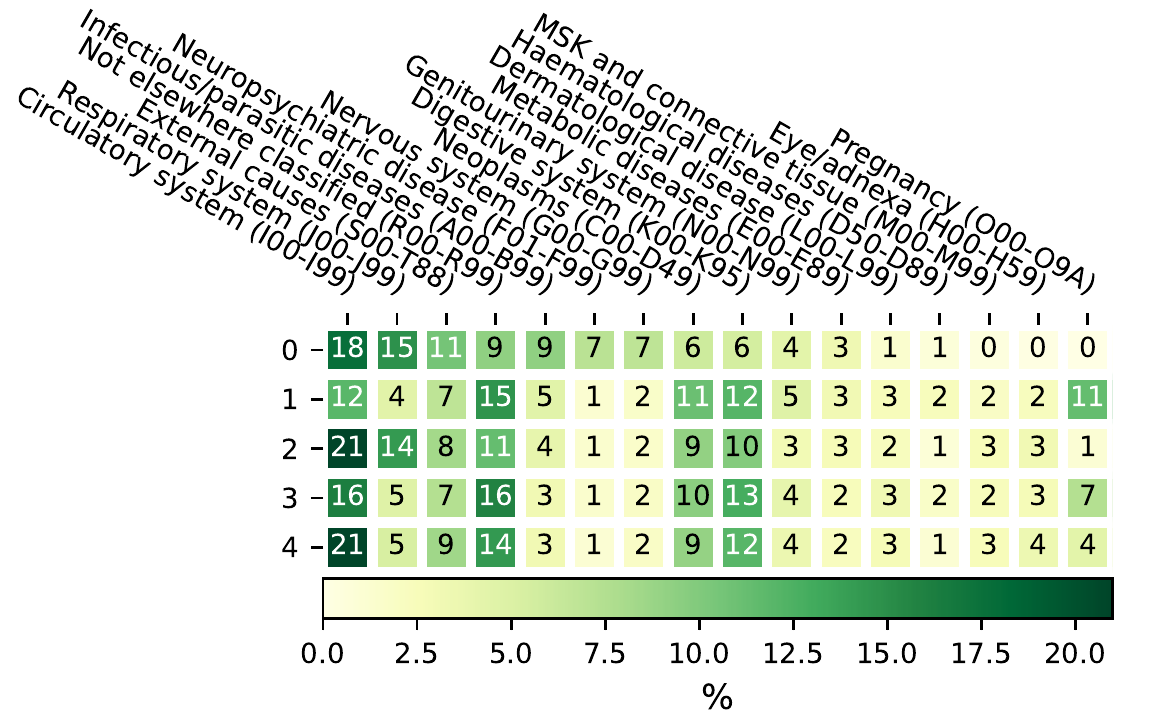}
    \caption{Heatmap of primary ICD10 codes as recorded by clinicians at the time of patient admission and collated by top-level grouping. For display purposes only ICD10 codes with $\geq$2\% incidence for at least one cluster are displayed. Since only a subset of ICD10 codes are visualised, each row does not add up to 100. MSK = Musculoskeletal.}
    \centering
    \label{Heatmap_ICD10codes}
\end{figure*}

\subsection{ICD10 Codes}
Figure \ref{Heatmap_ICD10codes} shows the frequency of identified primary 10th revision of International Classification of Diseases (ICD10) code groupings per cluster. The highest incidence ICD10 code group is `Circulatory system (I00-I99)' with 20.9\% and 21.0\% in clusters 2 and 4, respectively. Diseases of circulatory system are overall the most prevalent group. Diseases of the respiratory system are common in cluster 0 (14.8\%) and 2 (14.2\%), with approximately three times higher occurrence than in the other clusters. `Infectious/parasitic diseases (A00-B99)', `Neuropsychiatric disease (F01-F99)' and `Nervous system (G00-G99)' appear predominantly in cluster 0, whereas cases of `Neoplasms (C00-D49)' and `Digestive system (K00-K95)' are more common in the other clusters, ranging from 9.1\% to 12.8\% of all cases. Pregnancy-related incidences are most common in cluster 1 with 11.4\%.

\subsection{Surrogate Explanations}
\begin{figure*}[h]
    \centering
    \includegraphics[width=\textwidth]  
    {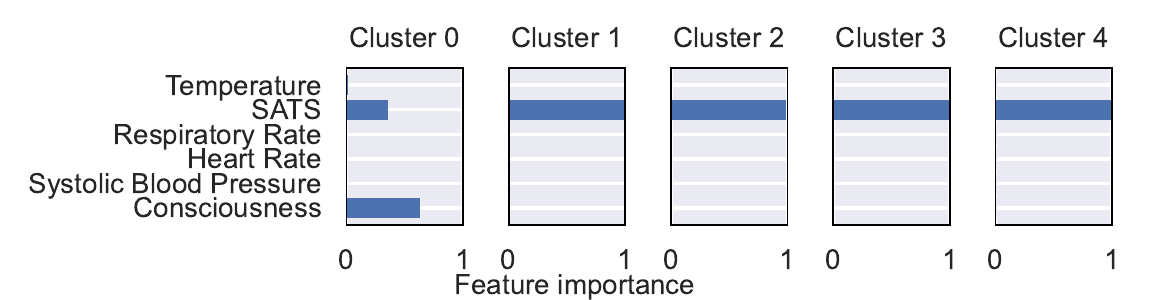}
    \caption{Surrogate explanations for the contribution of each vital in determining the assigment of patients into each cluster. SATS = Hemoglobin saturation with oxygen.}
    \label{Explanation}
\end{figure*}

Surrogate explanations suggest that the most important feature in determining cluster 0 assignment is consciousness (0.63 feature importance), followed by SATS (Figure \ref{Explanation}). For all other clusters, SATS is the most dominant factor, with a feature importance factor approaching 1.0. Temperature had a marginal influence for Clusters 0 and 2.

\subsection{Clinical Interpretation of Clusters}
The clinicians were able to detect inter-cluster differences and intra-cluster similarities. The limited level of consciousness and high NEWS in cluster 0, and different SATS levels for clusters 1, 2, 3, and 4 were identified as the main features. This mirrors findings of the vitals importance gained from the surrogate explanations.

Cluster 0 was found to have the highest NEWS. Yet, only one patient in the sampling group died. Cluster 1 was identified as a heterogenous group with SATS of 100\% and a low respiratory rate. Unsurprisingly, none of the analysed patients were admitted due to respiratory disease. That is in contrast to cluster 2 which showed high level of infection and/or respiratory disease. Based on the provided sample patients, this cluster appeared to have the highest mortality rate. SATS of 99\% and a low heart rate was the most characteristic feature of cluster 3. And patients in cluster 4 tended to have minimal background with medical conditions, low respiratory rate, low NEWS and SATS of 98\%. Apart from cluster 1, all clusters appeared to contain patients that did not match the cluster tendency.

\section{Discussion and Conclusion}
This study presents a pipeline to identify, explain and evaluate clusters of patient subtypes. 
Patient subtyping by way of clustering could be the first step towards a personalised scoring system, improving the predictive success of currently deployed risk scoring metrics \cite{rothman_development_2013,subbe_validation_2001,balkan_evaluating_2018,khwannimit_comparison_2007,abbott_pre-hospital_2018}.

The clusters identified in this study were based on just six vitals from the first set of readings taken during a hospital stay. Using only six vitals and including hospital departments beyond the ICU are in contrast to previous studies. Forte et al. \cite{castela_forte_identifying_2021} and Vranas et al. \cite{vranas_identifying_2017} included 76 and 23 clinical features, respectively, resulting in the identification of six subtypes of ICU patients. Here, cluster 0 appeared to predominantly include high-risk patients with a high NEWS and high level of limited consciousness, which was reflected in their elevated length of stay. The inter-cluster differences of the other clusters are harder to identify. Cluster 2 is the largest cluster with the second oldest population. Yet, the most frequent ICD10 code is equal in cluster 0 and 4. This indicates that patients admitted with the same condition may have different needs \cite{vranas_identifying_2017}. The clear, unique characteristics of cluster 0 and the stronger similarities of the other clusters are also reflected in the clinicians' feedback. They were able to identify SATS and level of consciousness as the key features of inter-cluster differences and successfully linked the cluster characteristics to the prevalent ICD10 codes. However, they also pointed out patients that did not match the cluster tendencies. This could be the result of the small sample size relative to the cluster size, distorting the clinicians' perception of the cluster characteristics. Additionally, after reading the first draft, the clinicians pointed out that the selected sample patients had a significant impact on which elements they considered as cluster characteristics. 

For further insights, the identified clusters were characterised and analysed regarding the occurrence of the most frequent ICD10 codes. Vranas et al \cite{vranas_identifying_2017} found `Sepsis' as the most common diagnosis in ICU patients in five out of six clusters, whereas Castela Forte et al \cite{castela_forte_identifying_2021} determined a different leading cause for each cluster. Here, circulatory diseases have the highest impact on hospital admissions, followed by `Not elsewhere classified' cases. Diseases of the Respiratory system appear to distinguish clusters 0 and 2 from the other clusters. \cite{castela_forte_identifying_2021} also identified two clusters with high prevalence of respiratory failure. 

Surrogate explainers were generated to improve cluster explainability. While consciousness is a key criterion for separating cluster 0 from the other clusters, SATS is the most dominant factor for distinguishing clusters 1-4. The level of consciousness has previously been identified as the key feature in predicting discharge from ICU \cite{mcwilliams_towards_2019}. The integration of surrogate explainers and clinicians helped validate and verify the presented results. Future studies and the deployment in hospital settings should consider this approach to increase fairness, accountability, and transparency. This also aids in building trust between the clinicians and machine learning systems. However, the identified patient subtypes should be treated with care as the whole analysis is based on a dataset from one hospital. Adding data from other hospitals as well as adding other features may reveal other or alter the identified patient subtypes. Furthermore, two clinicians were part of the team in order to co-design the process. Future studies will increase the number of clinicians for additional feedback, increasing the acceptance of and trust in the identified patient subtypes.

\section*{Acknowledgments}
This work was funded by Health Data Research UK via the Better Care Partnership Southwest (HDR CF0129). JC, AH and RSR are funded by the UKRI Turing AI Fellowship EP/V024817/1.

\bibliographystyle{spmpsci}
\bibliography{main}

\end{document}